\pdfoutput=1

\documentclass[11pt]{article}

\usepackage[]{acl}
\usepackage{times}
\usepackage{latexsym}

\usepackage{amsmath}
\usepackage{amsfonts}
\usepackage{cleveref}
\usepackage{booktabs}
\usepackage{subcaption}
\usepackage{xcolor}
\usepackage{tipa}

\crefname{section}{\S}{\S\S}
\crefname{table}{Tab.}{}
\crefname{figure}{Fig.}{}
\crefname{algorithm}{Algorithm}{}
\crefname{equation}{Eq.}{Eq.}
\crefname{appendix}{App.}{}
\crefname{theorem}{Theorem}{}
\crefname{prop}{Proposition}{}
\crefname{cor}{Corollary}{}

\definecolor{teal}{RGB}{26, 139, 140}
\definecolor{purple}{RGB}{136, 17, 136}
\definecolor{orange}{RGB}{254, 166, 2}
\definecolor{red}{RGB}{255, 0, 0}

\usepackage{todonotes}

\usepackage{emoji}
\newcommand{\apple}{\emoji[twitter_emoji]{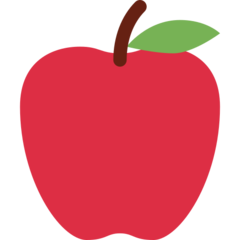}}
\newcommand{\ucambridge}{\emoji[twitter_emoji]{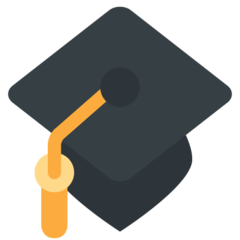}}

\usepackage{microtype}

\title{Prompting for a conversation: \\ How to control a dialog model?}

\author{Josef Valvoda$^{\ucambridge}$\thanks{\enspace Work done while at Apple.} \quad Yimai Fang$^{\apple}$ \quad David Vandyke$^{\apple}$ \\
  $^{\ucambridge}$University of Cambridge
  $^{\apple}$Apple \\
  \texttt{\href{mailto:jv406@cam.ac.uk}{jv406@cam.ac.uk}} \\
  \texttt{\{\href{mailto:yimai_fang@apple.com}{yimai\_fang},\href{mailto:dvandyke@apple.com}{dvandyke}\}@apple.com} \\
}

\date{}

\begin{document}
\maketitle
\begin{abstract}

Dialog modelling faces a difficult trade-off.
Models are trained on a large amount of text, yet their responses need to be limited to a desired scope and style of a dialog agent.
Because the datasets used to achieve the former contain language that is not compatible with the latter, pre-trained dialog models are fine-tuned on smaller curated datasets.
However, the fine-tuning process robs them of the ability to produce diverse responses, eventually reducing them to \emph{dull} conversation partners.
In this paper we investigate if prompting can mitigate the above trade-off.
Specifically, we experiment with conditioning the prompt on the query, rather than training a single prompt for all queries.
By following the intuition that freezing the pre-trained language model will conserve its expressivity, we find that compared to fine-tuning, prompting can achieve a higher BLEU score and substantially improve the diversity and novelty of the responses.
\looseness -1
\end{abstract}

\section{Introduction}

Prompting large language models (LLM) has recently demonstrated an impressive performance on a number of natural language processing (NLP) tasks such as machine translation  \cite{radford}, summarisation \cite{li-liang-2021-prefix} or question answering \cite{schick-schutze-2021-exploiting}.
Prompts are tokens which are appended or prepended to the input of a language model. 
They are employed to induce the model into generating useful information, while keeping the model weights frozen.
Soft prompts, continuous trainable vectors prepended to the model input, have in particular proven useful for a number of tasks \cite{liu2021pretrain}.
While requiring fine-tuning of only a relatively small number of parameters, they excel in a few-shot setting.
Furthermore, as the underlying LLM's grow in parameter size, they become competitive even in the full data setting \cite{lester-etal-2021-power}.
Simultaneously, prompts remove the burden of storing the full copy of a fine-tuned LLM for every task, which becomes increasingly useful as the LLM size grows.
Crucially for us, prompting preserves the LLM parameters which should help retain their general language abilities for a downstream task.
\looseness -1

\begin{figure}[t]
\includegraphics[width=\columnwidth]{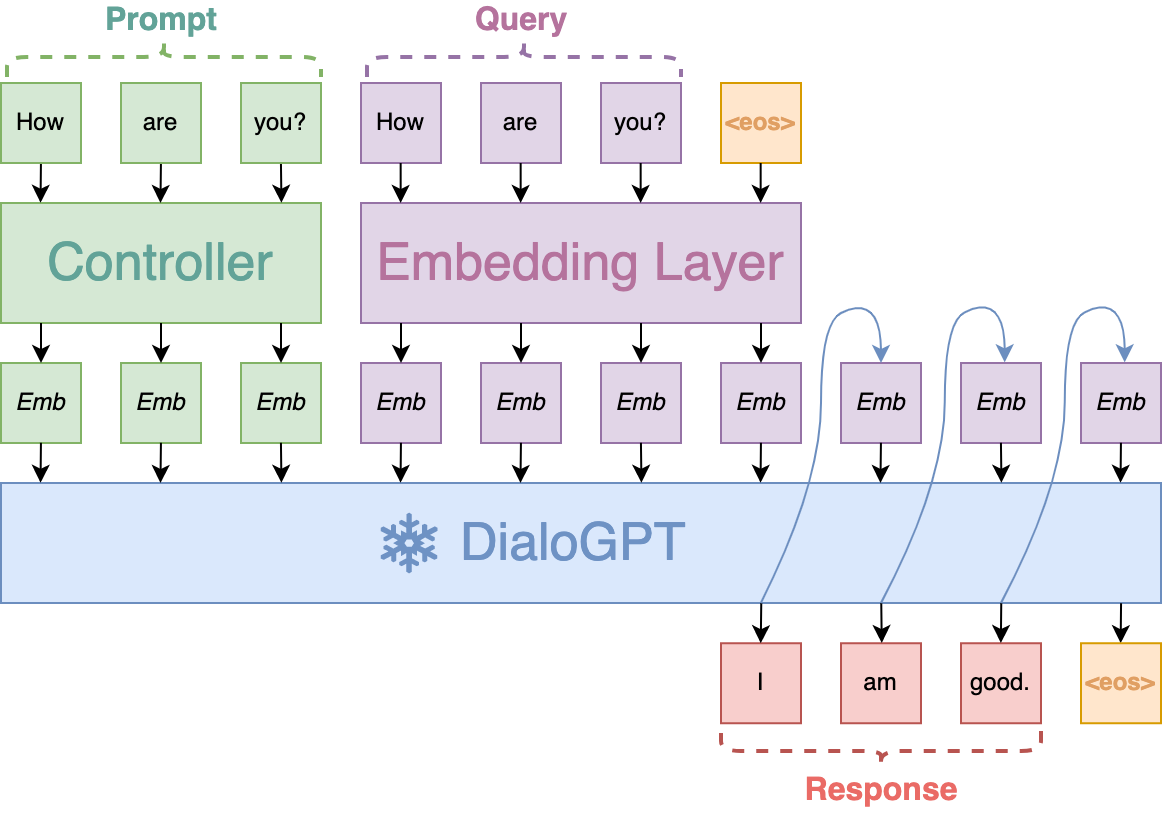}
\caption{Dynamic prompt conditions the \emph{prompt} on the \emph{query} using the \emph{controller}; a Transformer encoder.}
\label{fig:model}
\end{figure}


Dialog modelling is the task of generating a response given the previous dialog turn, the query \cite{li-etal-2016-deep}.
Dialog models are typically trained using the maximum-likelihood estimation (MLE) objective. However, MLE-trained models have a high propensity to provide dull responses, such as \emph{``I don't know''} \cite{sordoni-etal-2015-neural, serban, zhao-etal-2017-learning}. 
While state-of-the-art models, such as DialoGPT \cite{zhang-etal-2020-dialogpt}, can overcome this issue by training large models on massive amounts of data, a trade-off emerges. On the one hand, these models are expressive by virtue of the large datasets they are trained on. On the other hand, the same scale of training data and model parameters is responsible for the lack of control over the content of their responses. \looseness -1



Since dialog models are not useful without the ability to control their responses, in this paper, we turn to prompting as a possible method of exerting such control. Instead of fine-tuning the entire model, we keep the weights of the model intact, tuning only the prompts (and word embeddings) in an effort to preserve the models' expressivity. Furthermore, we develop a \textbf{dynamic-prompt},\footnote{Our work is concurrent to \citet{Gu2021} who develop a similar prompting method. However, our research is motivated by a different question from theirs. Specifically, we investigate how prompting can help with inducing creative responses, which we measure on the novelty and diversity metrics introduced below. Their work, on the other hand, focuses on improving performance on traditional metrics such as BLEU, NIST, METEOR and ROUGE-L.} which conditions the prompt on the query in an effort to dynamically induce the response by having a different prompt for every turn of the conversation, see \cref{fig:model}.
The intuition is to separate the task of language generation, which LLM's are very good at, from the task of selecting appropriate responses, which LLM's struggle to learn from the limited examples they are fine-tuned on.
\looseness -1

In our experiments with DailyDialog dataset, we find that the dynamic prompt outperforms both fine-tuning and the soft-prompting of DialoGPT in terms of a BLEU4 score. The best dynamic prompt model achieves $0.12$ BLEU4 compared to $0.11$ and $0.08$ of fine-tuning and soft-prompting respectively. Furthermore, the dynamic prompt finds the best trade-off between the BLEU score and novelty as well as the diversity of responses, maintaining above $0.90$ novelty and diversity as the BLEU4 score increases, while the other models can not.
Finally, we find that dynamic-prompting on GPT-2 achieves the best BLEU4 results, improving over prompted DialoGPT by an additional $19$\% and casting doubt about the utility of dialog-specific pre-training when it comes to prompting.
\looseness -1

\begin{figure*}
     \centering
     \begin{subfigure}{0.325\textwidth}
        \centering
        \includegraphics[width=\columnwidth]{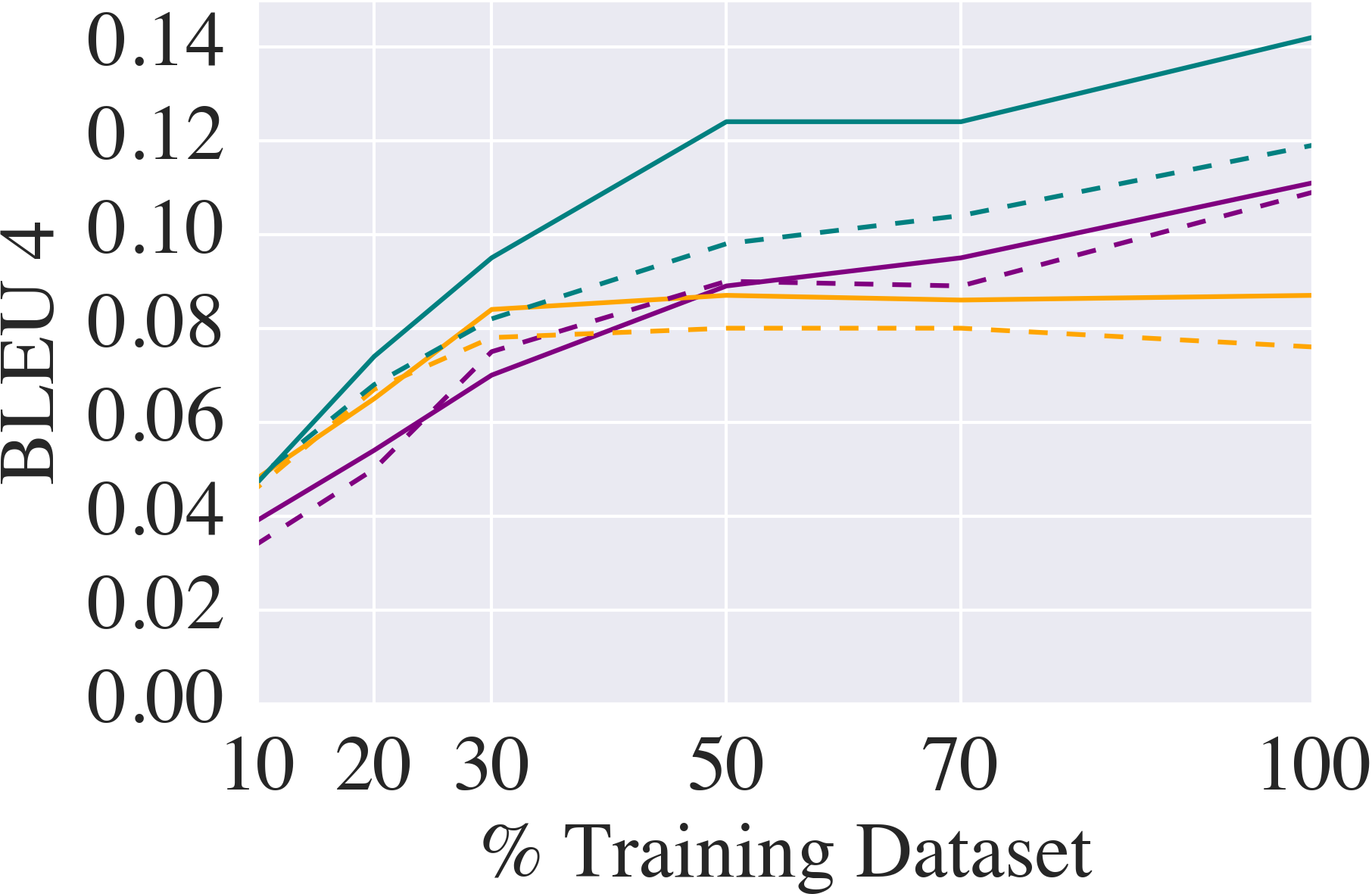}
     \end{subfigure}
     \hfill
     \begin{subfigure}{0.325\textwidth}
        \centering
        \includegraphics[width=\columnwidth]{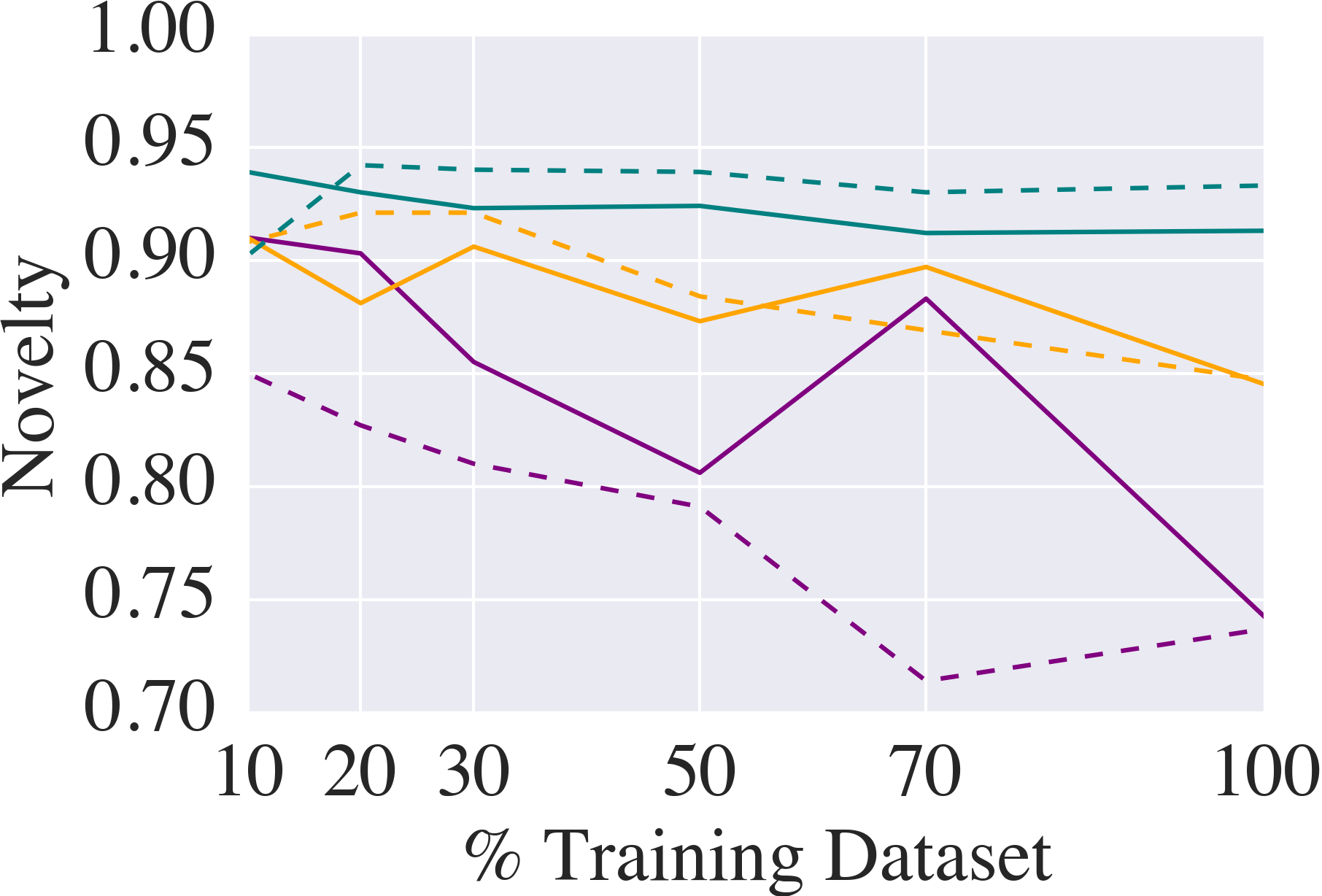}
    \end{subfigure}
    \hfill
     \begin{subfigure}{0.325\textwidth}
        \centering
        \includegraphics[width=\columnwidth]{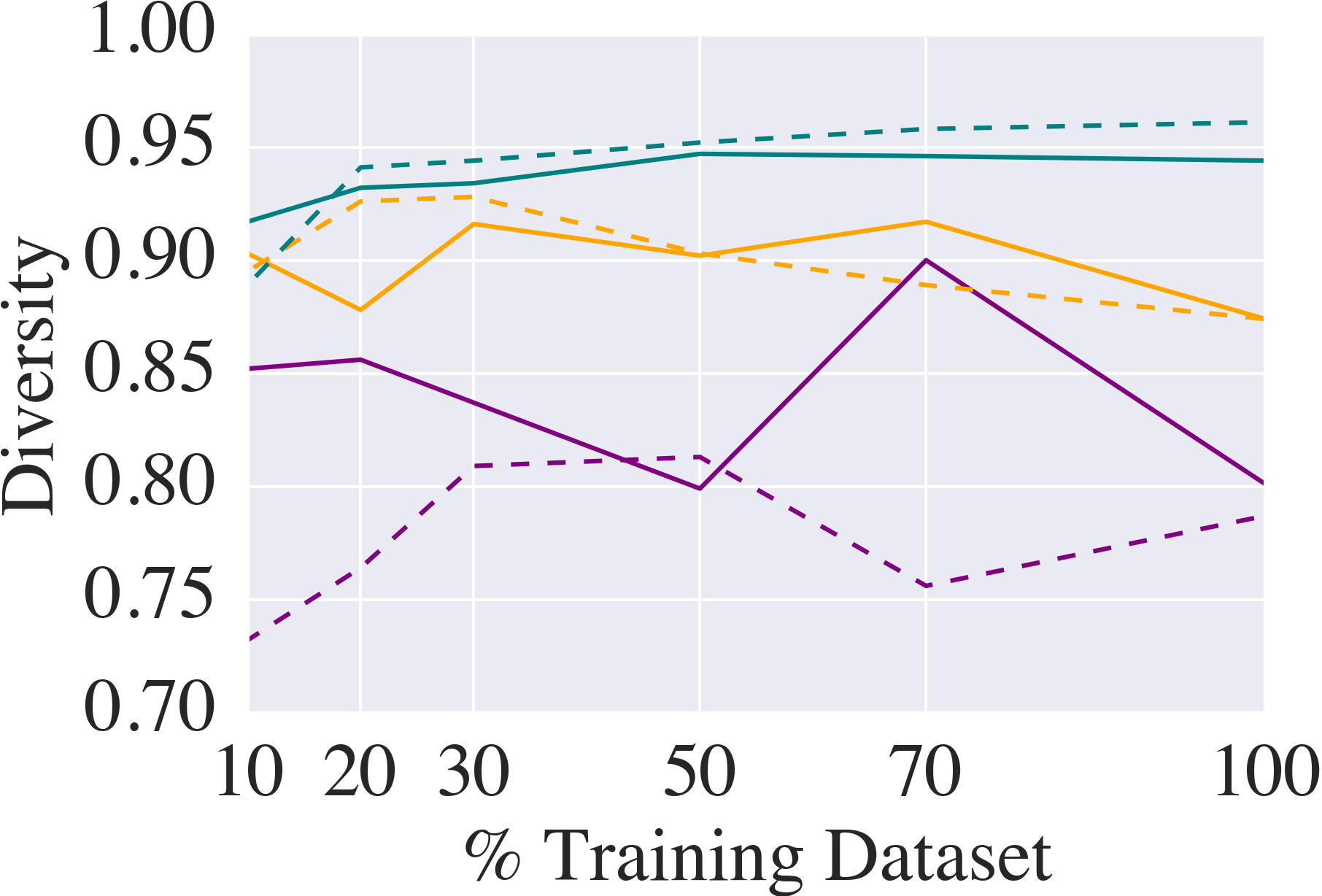}
     \end{subfigure}
        \caption{BLEU4, novelty and diversity scores for \textcolor{purple}{\textbf{fine-tuning}}, \textcolor{orange}{\textbf{soft-prompting}} and \textcolor{teal}{\textbf{dynamic-prompting}}. Full line is GPT-2, dashed line is DialoGPT.\looseness-1}
        \label{fig:results}
\end{figure*}

\section{Prompting for Dialog Generation}

Fine-tuning a pre-trained model on a small dialog dataset degrades the novelty of the responses \cite{sordoni-etal-2015-neural}. Prompting opens up a possibility of exerting control over the model behaviour, without touching the majority of the weights that might be useful for generating novel responses \cite{lester-etal-2021-power}.
However, typical prompting methods are restrained in using a single prompt for a single task.
In this section, we motivate the dynamic prompt as a natural next step in generalising the prompting paradigm for dialog modelling.
\looseness-1

\paragraph{Prompting.} Auto-regressive models with billions of parameters trained on a language modelling objective, such as OpenAI's GPT-3 \cite{NEURIPS2020_1457c0d6}, have demonstrated a strong few-shot performance without the need to update the model parameters. Instead of fine-tuning the model with target task examples, a manually designed prompt, for example in the form of a natural language sentence describing the task, is fed to the model to solicit the desired response.
For instance, to induce a translation the model might be told to: \emph{Translate English to French}. The description is followed up with pairs of examples of English sentences and their French translations. To derive a new translation the model is fed an English sentence alone and it is left to infer the French translation.
However, not all human-designed prompts elicit the desired response. In practice ensembles of many prompts have gone some way in improving the performance, but still necessitate humans in the loop to design the said prompts \cite{schick-schutze-2021-exploiting}.

\paragraph{Soft-prompting.}
To address this short-coming, recent work has found that prefixing and fine-tuning vocabulary tokens is a more expressive solution, which can learn the prompt from the data directly, without a need of a human prompt designer \cite{liu2021pretrain}. 
The soft prompts are not constrained by encoding existing tokens in the vocabulary and can freely encode parameters to facilitate the task.
Empirically, soft prompts achieve better performance than their \emph{hard} prompt counterparts and in a few-shot setting outperform fine-tuning of the entire model. \looseness -1

\paragraph{Dynamic-prompting.}
Soft prompts are restricted in utilising only a single prompt of a constant number of tokens for each task.
We hypothesise that there are many tasks where the desired response for every input will be hard to solicit through a single shared general prompt.
For example, the task of dialog generation has a general requirement to generate semantically and syntactically coherent, engaging responses.
However, an individual response might have specific properties that are only relevant within the context of the query.
While soft prompts are likely to solicit the more general tasks, such as machine translation or question answering, possibly because they exist in the training data to begin with, the more nuanced requirements might be heavily context-dependent.
Therefore, we condition the prompt on the context to enable learning different prompts for different queries.
We describe the method in detail below.

\section{Method}

\paragraph{Notation.} Let a dialogue be denoted as $x_{1} \cdots x_{N}$ where N is the sequence length. We denote dialogue history as $S = x_{1}, \cdots, x_{m}$ and target sequence as $T = x_{m+1}, \cdots, x_{N}$.
Now we can compute the conditional probability $P(T \mid S)$ as a product of a series of conditional probabilities:\looseness-1

\begin{equation}
    p(T \mid S)=\prod_{n=m+1}^{N} p\left(x_{n} \mid x_{1}, \cdots, x_{n-1}\right)
\end{equation}

This probability can be parametrized by an auto-regressive language model (in our case DialoGPT).




\paragraph{Fine-tuning.}
For our experimental baseline, we simply fine-tune pre-trained DialoGPT on the DailyDialog dataset, see \cref{sec:setup}.

\paragraph{Soft-prompting.} 
For soft-prompting, we prepend a prompt $F = z_{1},\cdots,z_{m}$ at the beginning of $S$. Prompt tokens are randomly initialised word embeddings appended to the DialoGPT vocabulary. 
In our experiments, we set the length of the soft prompt to the length of $S$.
Now, to compute the probability of the target given the dialogue history and prompt, we compute:
\begin{align}
&p(T \mid S, F)= \\ \nonumber
&\prod_{n=m+1}^{N} p\left(x_{n} \mid z_{1}, \cdots, z_{m}, x_{1}, \cdots, x_{n-1}\right)
\end{align}
We instantiate the pre-trained DiaoGPT, but this time we freeze all of the model parameters, except the word embedding weights. 
While in a typical soft-prompting experiment only the prompt embeddings are tuned, we found that for our dialog setting the performance suffers considerably when the model is constrained to train only the parameters of prompt embeddings, which is why we relax this requirement. 
Furthermore, we do not calculate the loss for the logits corresponding to the prompt. 
This is because we don't know the ground truth of what the prompt should be.



\paragraph{Dynamic-prompting.}
Finally, we extend the soft-prompting paradigm by conditioning the prompt $F$ on the query $S$ to find a unique prompt for every query, see \cref{fig:model}. 
We use an auto-regressive Transformer encoder to generate the prompt embeddings: 
\begin{equation}
    \textbf{h}_{1} , \cdots , \textbf{h}_{m} = \mathrm{Transformer} ( x_{1}, \cdots ,x_{m} )
\end{equation}
Now, we use $\textbf{h}$ as our prompt token embeddings and prepend them to the query embeddings inside the DialoGPT model. The model is otherwise trained the same way as the soft-prompting model above. The Transformer encoder is trained jointly with the embeddings.\looseness-1

\section{Experimental Setup}\label{sec:setup}
\paragraph{Datasets.}
We conduct our experiments on the DailyDialog dataset consisting of $13,118$ dialogues, split into $11,118$/$1,000$/$1,000$ training/validation/test sets \cite{li-etal-2017-dailydialog}. We only focus on single-turn dialog modelling and process the dataset so that our pairs of queries and responses correspond to every two steps in a conversation.
Inspired by \citet{li-liang-2021-prefix}, we train our models on $10$\%, $20$\%, $30$\%, $50$\%, $70$\%, and $100$\% of the DailyDialog dataset to observe how the number of training samples affects the performance. 
We keep the validation and test sets full-sized for all experiments.
For full experimental details, see \cref{appendix:details}.
\looseness-1

\paragraph{Metrics.}
We follow \citet{li-etal-2017-dailydialog} and evaluate the models using BLEU4 score. Since we do not want the model to simply repeat the responses it has memorised from the training data, i.e. the dull response issue, we additionally introduce two new metrics: \textbf{novelty} and \textbf{diversity}. 
We define novelty as the proportion of model outputs on the test set that are not found in the training set.
Given a test set, diversity is defined as the number of unique model outputs divided by the total number of outputs.
A good dialog model should be able to achieve a high BLEU4 score, while maintaining a high level of novelty and diversity in its responses.
\looseness-1

\paragraph{Models.}
We experiment with two models, DialoGPT \cite{zhang-etal-2020-dialogpt}, a state-of-the-art dialog model, and GPT-2 \cite{radford} the model DialoGPT is fine-tuned from.
We choose the latter model to gain insight into how useful in-domain pre-training of DialoGPT is for prompting.\looseness-1

\section{Results}
Our main results are contained in \cref{fig:results}.
First, we compare the performance of prompting vs fine-tuning on the DialoGPT model.
We find that under the full data setting the dynamic prompt model outperforms the soft prompt model and even the fine-tuned model on all three metrics. 
With a BLEU4 of $0.12$ dynamic prompt is $9$\% better than fine-tuning ($0.11$) and $15$\% better than a soft prompt ($0.08$). 
The difference is even more dramatic when we compare the corresponding diversity and novelty scores. 
Dynamic prompt achieves a novelty of $0.93$ and diversity of $0.96$, an improvement of $9$\% ($0.85$) and $10$\% ($0.87$) respectively over the soft prompt, and an even more pronounced improvement of $26$\% ($0.74$) and $22$\% ($0.79$) respectively over fine-tuning the model.
\looseness-1

Next, we observe that soft-prompting is competitive in the low data setting and outperforms fine-tuning. 
However, with more than $30$\% of the training data available, soft-prompting stops improving altogether and begins to under-perform the other models.
Dynamic prompt on the other hand maintains the best BLEU4 score no matter the amount of training data.
\looseness -1

Now we turn to the question of novelty and diversity degradation. 
We observe that the dynamic prompt does not suffer from this issue nearly as much as the other models.
In contrast, for the soft-prompted and fine-tuned models, the better the BLEU4 score gets the lower the novelty drops.
While soft prompt already mitigates this drop-off, degrading slower than DialoGPT,
dynamic prompt does not suffer from this effect and maintains above $90$\% novelty throughout. 
Similarly, for diversity, the dynamic prompt maintains around $0.95$ score for any percentage of training data, while the soft prompt drops down over time and fine-tuning hovers around $0.80$.
\looseness-1

Finally, we can turn to the comparison of DialoGPT and GPT-2.
Against our expectations, DialoGPT under-performs GPT-2 on BLEU4 despite the former having been pre-trained on dialog-specific data.
The performance improvement is most notable in the case of the dynamic prompt, where GPT-2 ($0.14$) achieves $19$\% higher performance than DialoGPT ($0.12$).
On the other hand, fine-tuning either model leads to nearly identical performance.
You can find our discussion of this phenomenon in \cref{appendix:discussion}, all our results in \cref{appendix:results} and example outputs in \cref{appendix:examples}.
\looseness-1



\section{Related Work}
Our work builds on two strains of thought. First, we deal with the problem of dull responses in dialog modelling. Related work includes the use of reinforcement learning \cite{li-etal-2016-deep}, latent variables \cite{cao-clark-2017-latent} and decoding techniques to mitigate this issue \cite{LiMJ16}. \looseness -1

Second, we build on the idea of continuous prompting, which was developed concurrently by \citet{lester-etal-2021-power} and \citet{li-liang-2021-prefix}. There are many variations of the prompting paradigm. For instance, we fine-tune the embeddings along with the prompts, but \citet{liu2021gpt} tune the prompt with the full model. \citet{schick-schutze-2021-just} on the other hand tune only the model, while keeping the prompt fixed.
\looseness -1


\section{Conclusion}
We find dynamic-prompting bests fine-tuning by generating more novel and diverse responses with a higher BLEU score while training only a small portion of the DialoGPT/GPT-2 parameters.
Thus proving itself as a useful method for mitigating dialog modelling issue of \emph{dull} responses.
Prompting the general purpose GPT-2 achieves much higher performance than prompting the specialist DialoGPT model, suggesting that for pre-training data, diversity is more valuable for dialog modelling than dialog-specific information.
\looseness-1

\bibliography{anthology,acl2020,other}
\bibliographystyle{acl_natbib}

\appendix
\section{Appendix A: Discussion}\label{appendix:discussion}

\citet{liu2021pretrain} designed their soft prompt based on the observation that context can control the LLM without the need to change its parameters.
They demonstrate that the context might be hard to find among the existing word embeddings and instead learn it via back-propagation.
We go a step further and demonstrate that when it comes to dialog generation it is harder to learn a single prompt than to learn a function that conditions the prompt on the query.
We believe that this is because there is no single context, represented by words or embeddings, that can capture a complex task such as a conversation.
This becomes intuitive if we separate the task of language generation from the task of making a conversation.
The conversation is an interaction of the query with the state of the conversation agent.
Without the ability to respond to the query the agent/language model is simply saying whatever is the most probable response.
A prompt can induce some general modification over this response, but there is no process by which the agent can react to the query.
In our dynamic prompt setting, the agent, represented by the controller module, is allowed to learn how to respond to each query. 
\looseness -1

The separation of thought from language production is of course not a new idea.
Under the language of thought hypothesis, \emph{mentalese} is the language of thought \cite{sep-language-thought}.
While similar to natural language in its compositional structure, mentalese is separate from language itself.
From this perspective, building models that separate the task of learning a language from that of learning how to use language makes perfect sense.
Following this intuition, it also makes sense to pre-train the language model on general language, rather than a specific task.
We believe that in our setting, the superior language ability of GPT-2 allows for a better Controller in the dynamic prompt setting.

\section{Appendix B: Experimental Details}\label{appendix:details}
We implement our models using the Pytorch and Huggingface libraries.
We experiment with total of $36$ configurations, testing three learning methods (Fine-tuning, Soft-prompting and Dynamic-prompting) on two LLMs (DialoGPT and GPT-2) and six data regimes ($10$\%, $20$\%, $30$\%, $50$\%, $70$\%, and $100$\%).
In each configuration, we run $12$ trials of different learning rates $\in [3\times10^{-6}, 0.009]$, with a batch size of $8$, for a maximum of $300$ epochs (with early stopping after $100$ epochs), and select the best model by validation BLEU.
Each trial is done on a single GPU with $32$GB memory, and the maximum training time is $14$ days. To generate the model output, we always use only greedy decoding.
\looseness-1
\section{Appendix C: Full Results}\label{appendix:results}

\begin{figure}[h]
     \centering
     \begin{subfigure}{\columnwidth}
        \centering
        \includegraphics[width=\columnwidth]{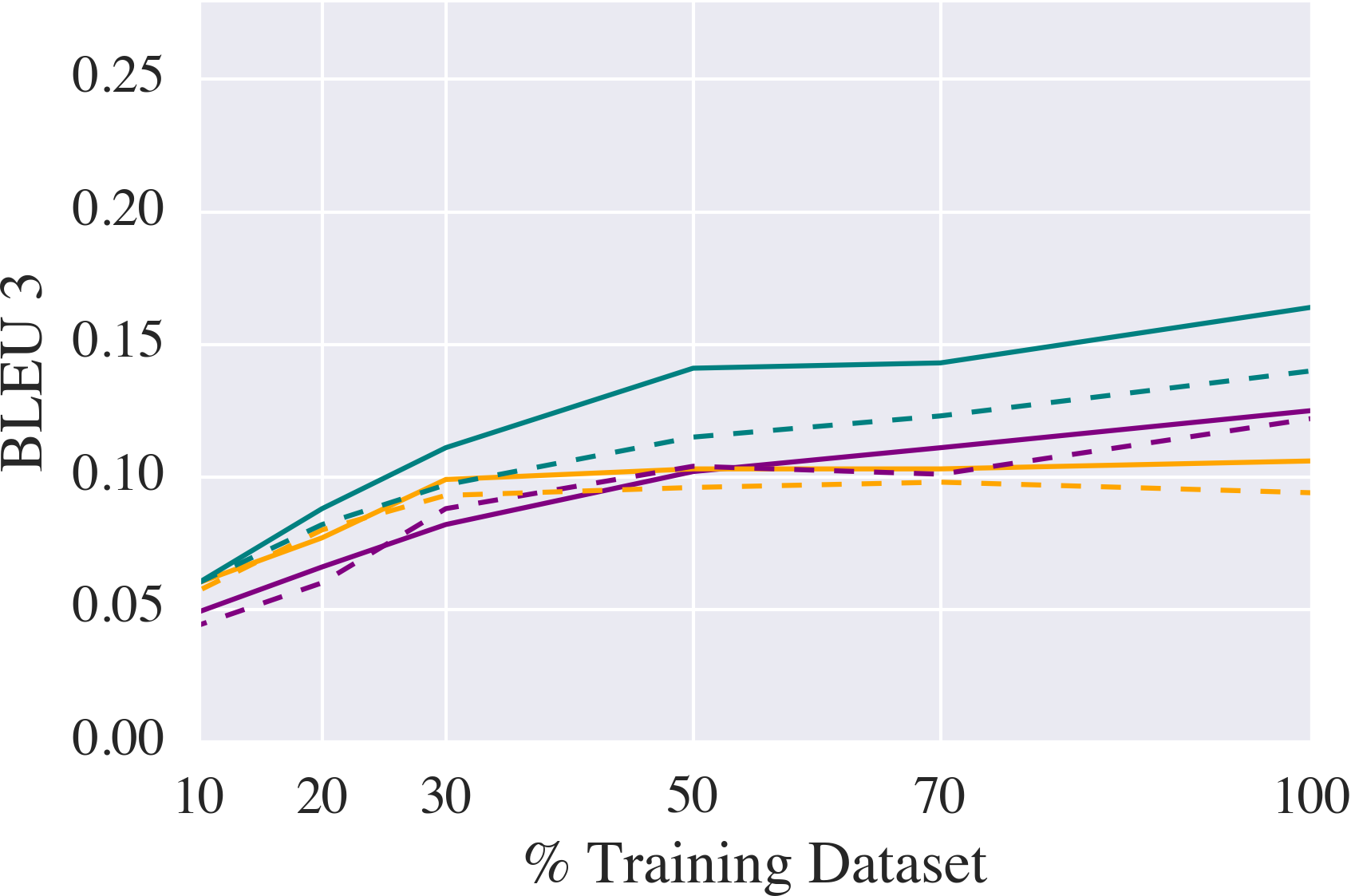}
     \end{subfigure}
     \hfill
     \begin{subfigure}{\columnwidth}
        \centering
        \includegraphics[width=\columnwidth]{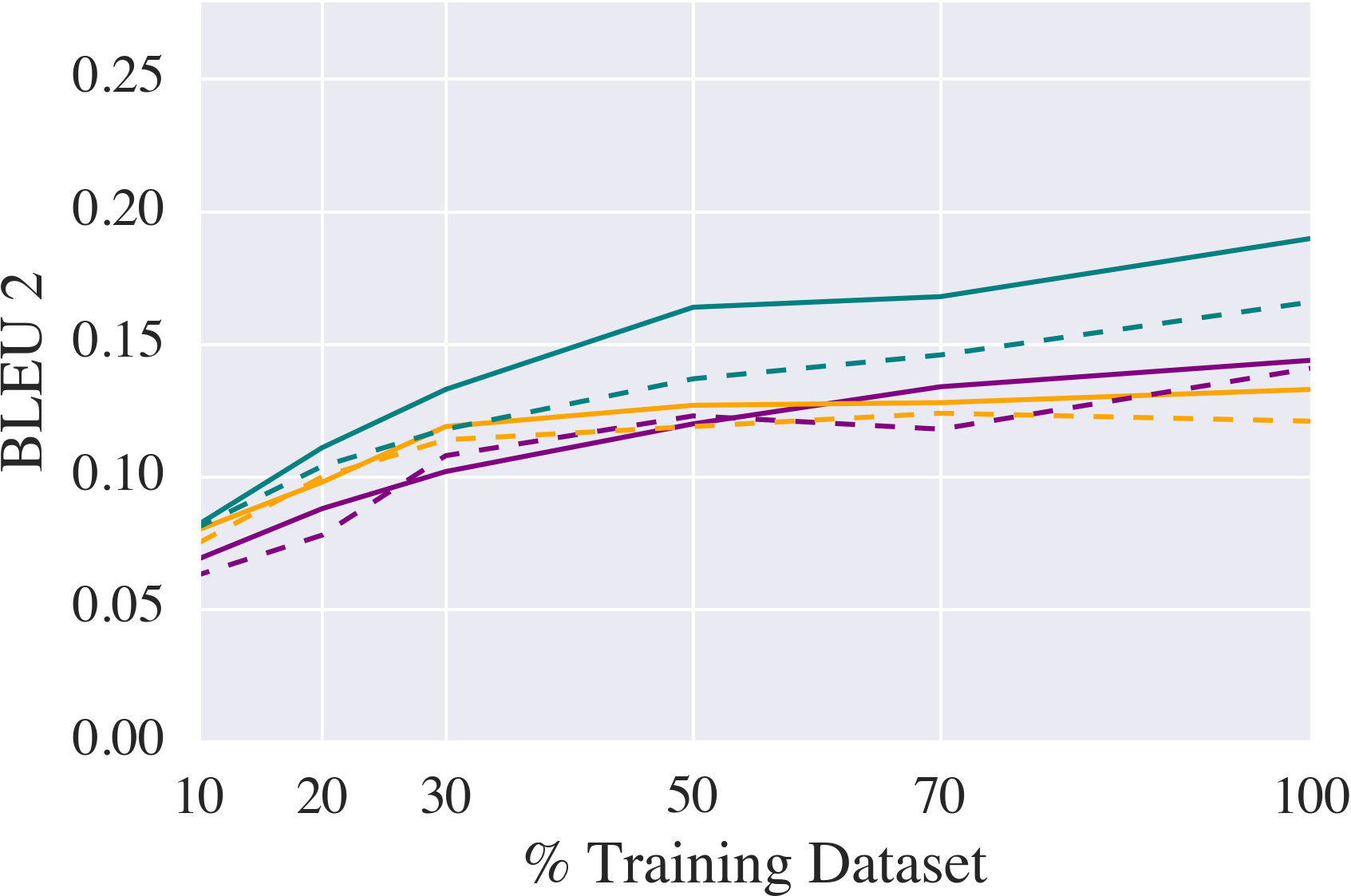}
    \end{subfigure}
    \hfill
     \begin{subfigure}{\columnwidth}
        \centering
        \includegraphics[width=\columnwidth]{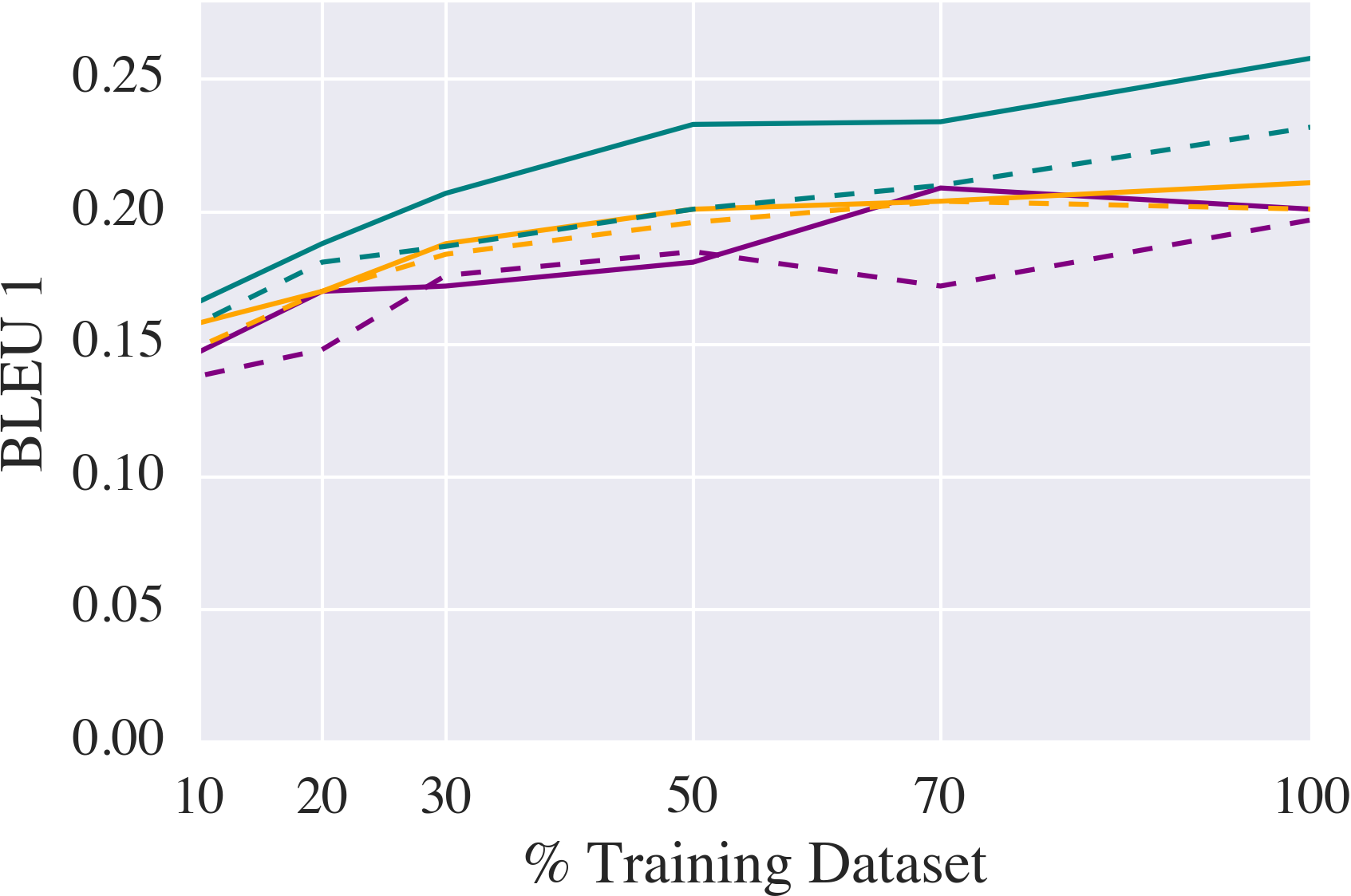}
     \end{subfigure}
        \caption{BLEU scores for \textcolor{purple}{\textbf{fine-tuning}}, \textcolor{orange}{\textbf{soft-prompting}} and \textcolor{teal}{\textbf{dynamic-prompting}}. Full line is GPT-2, dashed line is DialoGPT.\looseness-1}
        \label{fig:results}
\end{figure}

\begin{table*}[hb]
\centering
\begin{tabular}{llrrrrrr}
\toprule
        Data &   Model & BLEU1 &  BLEU2 &  BLEU3 &  BLEU4 &  Novelty & Diversity\\
\midrule
  10\%  &    fine-tuning &  0.138 &  0.063 &  0.044 &  0.034 &     0.850 &       0.732 \\
             &    soft-prompting &  0.149 &  0.075 &  0.057 &  0.046 &     0.908 &       0.895 \\
             & dynamic-prompting &  0.158 &  0.081 &  0.060 &  0.048 &     0.902 &       0.889 \\
             \midrule
         20\% &    fine-tuning &  0.148 &  0.078 &  0.060 &  0.050 &     0.827 &       0.764 \\
             &    soft-prompting &  0.170 &  0.100 &  0.080 &  0.067 &     0.921 &       0.926 \\
             & dynamic-prompting &  0.181 &  0.104 &  0.082 &  0.068 &     0.942 &       0.941 \\
             \midrule
         30\% &    fine-tuning &  0.176 &  0.108 &  0.088 &  0.075 &     0.810 &       0.809 \\
             &    soft-prompting &  0.184 &  0.114 &  0.093 &  0.078 &     0.921 &       0.928 \\
             & dynamic-prompting &  0.187 &  0.118 &  0.097 &  0.082 &     0.940 &       0.944 \\
             \midrule
         50\% &    fine-tuning &  0.185 &  0.123 &  0.104 &  0.090 &     0.791 &       0.813 \\
             &    soft-prompting &  0.196 &  0.119 &  0.096 &  0.080 &     0.884 &       0.903 \\
             & dynamic-prompting &  0.201 &  0.137 &  0.115 &  0.098 &     0.939 &       0.952 \\
             \midrule
         70\% &    fine-tuning &  0.172 &  0.118 &  0.101 &  0.089 &     0.714 &       0.756 \\
             &    soft-prompting &  0.204 &  0.124 &  0.098 &  0.080 &     0.869 &       0.889 \\
             & dynamic-prompting &  0.210 &  0.146 &  0.123 &  0.104 &     0.930 &       0.958 \\
             \midrule
100\%  &    fine-tuning &  0.197 &  0.141 &  0.122 &  0.109 &     0.737 &       0.787 \\
             &    soft-prompting &  0.201 &  0.121 &  0.094 &  0.076 &     0.847 &       0.874 \\
             & dynamic-prompting &  0.232 &  0.166 &  0.140 &  0.119 &     0.933 &       0.961 \\
\bottomrule
\end{tabular}
\caption{The full results of our experiments on DailyDialogue with DialoGPT.}
\label{tab:results}
\end{table*}

\begin{table*}[hb]
\centering
\begin{tabular}{llrrrrrr}
\toprule
         Data &   Model & BLEU1 &  BLEU2 &  BLEU3 &  BLEU4 &  Novelty & Diversity\\
\midrule
  10\%  &    fine-tuning &  0.147 &  0.069 &  0.049 &  0.039 &     0.910 &       0.852 \\
             &    soft-prompting &  0.158 &  0.080 &  0.060 &  0.048 &     0.910 &       0.903 \\
             & dynamic-prompting &  0.166 &  0.082 &  0.060 &  0.047 &     0.939 &       0.917 \\
         20\% &    fine-tuning &  0.170 &  0.088 &  0.066 &  0.054 &     0.903 &       0.856 \\
             &    soft-prompting &  0.170 &  0.098 &  0.077 &  0.065 &     0.881 &       0.878 \\
             & dynamic-prompting &  0.188 &  0.111 &  0.088 &  0.074 &     0.930 &       0.932 \\
         30\% &    fine-tuning &  0.172 &  0.102 &  0.082 &  0.070 &     0.855 &       0.837 \\
             &    soft-prompting &  0.188 &  0.119 &  0.099 &  0.084 &     0.906 &       0.916 \\
             & dynamic-prompting &  0.207 &  0.133 &  0.111 &  0.095 &     0.923 &       0.934 \\
         50\% &    fine-tuning &  0.181 &  0.120 &  0.102 &  0.089 &     0.806 &       0.799 \\
             &    soft-prompting &  0.201 &  0.127 &  0.103 &  0.087 &     0.873 &       0.902 \\
             & dynamic-prompting &  0.233 &  0.164 &  0.141 &  0.124 &     0.924 &       0.947 \\
         70\% &    fine-tuning &  0.209 &  0.134 &  0.111 &  0.095 &     0.883 &       0.900 \\
             &    soft-prompting &  0.204 &  0.128 &  0.103 &  0.086 &     0.897 &       0.917 \\
             & dynamic-prompting &  0.234 &  0.168 &  0.143 &  0.124 &     0.912 &       0.946 \\
100\%  &    fine-tuning &  0.201 &  0.144 &  0.125 &  0.111 &     0.742 &       0.801 \\
             &    soft-prompting &  0.211 &  0.133 &  0.106 &  0.087 &     0.845 &       0.874 \\
             & dynamic-prompting &  0.258 &  0.190 &  0.164 &  0.142 &     0.913 &       0.944 \\
\bottomrule
\end{tabular}
\caption{The full results of our experiments on DailyDialogue with GPT-2.}
\end{table*}

\onecolumn
\section{Appendix D: Examples}\label{appendix:examples}

\begin{table*}[hb]
    \centering
    \begin{tabular}{p{0.22\linewidth}  p{0.22\linewidth}  p{0.22\linewidth}  p{0.22\linewidth}}
    \toprule
      \textbf{Query} & \textcolor{teal}{\textbf{Dynamic-prompting}} & \textcolor{purple}{\textbf{Fine-tuning}} & \textcolor{orange}{\textbf{Soft-prompting}} \\ \midrule
      I think it unwise for either of us to be inflexible. How about meeting each other halfway?
      &  
      Do I know how to meet these employees?
      & 
      \textcolor{red}{Then, what's your countr-offer?} 
      & 
      That's verywise for either of us. How about meeting another person in person?
      \\\\
      
      It's a success all right. The kids keep trying to catch me. 
      & 
      You better attach some good ones too. 
      & 
      I'm getting more and more special. 
      & 
      \textcolor{red}{I'm glad you're here.}
      \\\\
      
      You can audition music on line, and you can audition it off line after downloading it, so you needn't pay for it.
      &
      But when I am getting done I will make a big fan list.
      &
      Is that a lot of people's music here?
      &
      Actually, I have been thinking about taking the notes from the book.
      \\\\
      
      What date would you like to fly? Saturday, July 25. How many people will be traveling?
      &
      You'd better get a lot of tickets first.
      &
      A few people are coming to get me.
      &
      Two flight attendants will be coming with you.
      \\\\
      
      It's more violent than TV news! What time is it anyway?
      &
      I don't know, but you are still beating around the bush. You aren't going to be able to stop anybody who looks like you.
      &
      Oh, man! I didn't notice it.
      &
      It's after ten o'clock, after the candles, so it's hard for me to make a decision.
      \\
    \bottomrule
    \end{tabular}
    \caption{Example responses generated by the models under consideration. \textcolor{red}{Red} indicates reused training data response.}
    \label{tab:my_label}
\end{table*}

\end{document}